\documentclass{article}
\usepackage[legalpaper, margin=0.5in]{geometry}
\usepackage{blindtext}
\usepackage{float}
\usepackage{graphicx}
\usepackage{amssymb,amsmath}
\usepackage{multicol}
\title{Multi-agent Coverage Control: From Discrete Assignments to Continuous Multi-agent Distribution Matching }
\author{Solmaz S. Kia and Sonia Mart{\'\i}nez\\~\\
}
\date{}
\begin{document}
\maketitle

\begin{multicols}{3}

\section*{Abstract}
The multi-agent spatial coverage control problem encompasses a broad research domain, dealing with both dynamic and static deployment strategies, discrete-task assignments, and spatial distribution-matching deployment. Coverage control may involve the deployment of a finite number of agents or a continuum  through centralized or decentralized, locally-interacting schemes. 
All these problems can be solved via a different taxonomy of deployment algorithms for multiple agents. Depending on the application scenario, these problems involve from purely discrete descriptions of tasks (finite loads) and agents (finite resources), to a mixture of discrete and continuous elements, to fully continuous descriptions of the same. Yet, it is possible to find common features that underline all the above formulations, which we aim to illustrate here.  By doing so, we aim to point the reader to novel references related to these problems. 

The short article outline is the following: 

\begin{itemize}
    \item Static coverage via concurrent area partitioning and assignment.
    \item Static coverage as a discrete task assignment.
    \item Continuum task assignment for large-scale swarms.
\end{itemize}

\section{Introduction}

The coverage control problem concerns the strategic placement of a limited resource, such as sensors or robots (hereafter referred to as agents), across an area of interest to optimize a specific coverage measure. One of the earliest instances of such problems can be traced back to the work of the German astronomer and mathematician Johannes Kepler in $1611$. Known as the Kepler Conjecture, it proposed that the densest arrangement of equally-sized spheres in three-dimensional space is the face-centered cubic packing (or hexagonal close packing). While this primarily applies to spheres, it laid the groundwork for studying the packing of circles (disks) in two dimensions. The sphere/circle packing problem solution continues to be relevant and has been employed in various wireless sensor deployment problems (see, e.g.,~\cite{hales1998overview} for an overview and~\cite{lam2008sensor} and~\cite{mozaffari2016efficient} for applications). However, the problems faced in multi-agent coverage are often much more intricate, often involving non-homogeneous agents whose footprints are not necessarily isotropic uniform disks. The number of agents can also be significantly less than what is needed for full coverage. Moreover, not all areas hold equal importance, and agent deployment is expected to be aligned with some area priority measure. Additionally, the deployment objective is not always a static configuration and may include goals such as persistent monitoring or dynamic surveillance.

In scenarios with a limited number of agents, the primary strategy involves dividing the area into sub-regions and assigning an agent to each. Often, geographical statistical analysis or prior information about the event of interest is used to extract a spatial probability distribution of the event of interest to guide the deployment of the agents. This distribution can be derived from various sources, such as high-altitude imaging, satellite imagery, or historical data, and can also be dynamically adjusted through online learning. To explore these complexities and strategies further, in this article we will discuss various approaches to task assignment and agent deployment in different contexts.

The general problem setting that we consider consists of a multi-agent deployment problem, where the goal is to deploy a group of $N$ agents over a finite two-dimensional convex polytope  $\mathcal{W} \subset \mathbb{R}^2$ to provide a \emph{service}. The service can involve sensor deployment for data collection/harvesting or event detection, dispatch for service, or providing wireless hotspots. We let $\phi: \mathcal{W} \rightarrow [0,1]$ be an \emph{a priorily} known stationary spatial probability density function that serves as the area priority function. The function $\phi$ can describe various scenarios, such as the distribution of crowds or animals, information sources, pollution spills, or forest fires.

Although the general problem setting is common, once we introduce specific details such as the agents' service model, coverage objective, and operational requirements, different solution approaches must be employed to solve the problem effectively. It is important to note that finding a globally optimal coverage configuration for multi-agent problems is usually very difficult, with many related facility localization problems (e.g., p-center and p-median) being NP-hard~\cite{megiddo1984complexity}. As such, various heuristic and approximation methods are often necessary to achieve practical and scalable solutions.

The remainder of this article is arranged as follows: Section~\ref{sec::voronoi} discusses locational optimization for optimal agent deployment using techniques like Voronoi partitions and power diagrams, and how these frameworks address heterogeneous agents and anisotropic sensory systems. Section~\ref{sec::disc_assign} presents an alternative approach for multi-agent deployment by introducing a two-step procedure for identifying Points of Interest (PoIs) and solving the deployment problem as a discrete assignment problem. It explores methods such as Gaussian Mixture Models (GMM), $K$-means clustering, and the Stein Variational Gradient Descent (SVGD) method to identify and utilize PoIs effectively, followed by assignment using optimal bipartite matching and submodular maximization frameworks. Section~\ref{sec::continum} shifts focus to large-scale multi-agent systems, framing the problem at a macroscopic scale as a distribution matching problem. This section covers the use of the Wasserstein-Kantorovich metric for optimal transport problems and discusses algorithms for large-scale deployments that ensure microscopic constraints for sensing, communication, and control. Section~\ref{sec::conclude} summarizes the key points discussed in the article, emphasizes the significance of multi-agent spatial coverage problems in robotics, and outlines future research directions.

\begin{figure*}[t]
\centering
\includegraphics[width=0.7\textwidth]{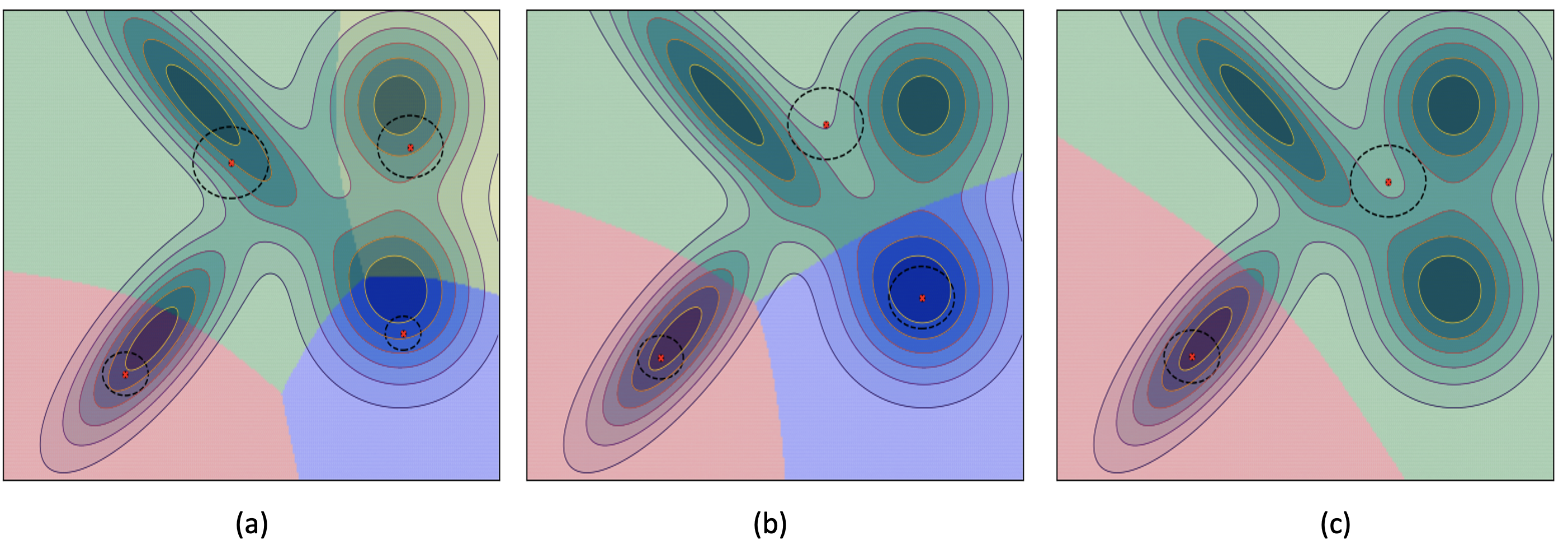}
\caption{Weighted Voronoi-based deployment based on minimizing~\eqref{eq::power_diagram}. The contour plot in the background shows the area priority density distribution function $\phi$. The agents final deployment positions shown by red dots and their power disks shown with dashed circles. Plots (a), (b) and (c) show respectively, deployment results for four, three and two heterogeneous agents. Figure
courtesy of Donipolo Ghimire.}\label{fig::voronoi}
\end{figure*}

\section{Static coverage via concurrent area partitioning and assignment}\label{sec::voronoi}
In a static multi-agent coverage scenario for a group of $N$ finite agents $\mathcal{A} = \{1, \cdots, N\}$, the objective is to solve a locational optimization problem, whose outcome determines the agents' final deployment locations $P = \{p_1, \cdots, p_N\} \subset \mathcal{W}$ such that some coverage objective tied to $\phi$ is optimized. This objective often depends on the the specific type of service the agents are intended to provide.

Let us consider a facility location problem where the objective is to ensure timely dispatch or fair access to the agents' service for points in the area $\mathcal{W}$, based on the area priority function $\phi$. The service quality of an agent at any point $q \in \mathcal{W}$, provided by the $i$-th agent deployed at location $p_i$, often degrades with distance. This degradation is typically modeled by a non-decreasing differentiable function $f(\|q - p_i\|): \mathbb{R}_{\geq 0} \to \mathbb{R}_{\geq 0}$. To achieve the deployment objective, we solve the following locational optimization formulation:
\begin{align}\label{eq::voronoi_opt}
  &  \min H(P, \bar{\mathcal{W}}), \\
 &     H(P, \bar{\mathcal{W}})= \sum_{i=1}^N\! \int_{\mathcal{W}_i} f(\|q - p_i\|) \text{d}(\phi(q)),\nonumber
\end{align} 
where $\mathcal{W}$ is partitioned into disjoint subsets $\bar{\mathcal{W}}=\{\mathcal{W}_1, \cdots, \mathcal{W}_N\}$, and $\mathcal{W} = \cup_{i=1}^N \mathcal{W}_i$, with each subset assigned to an agent as the agent's ``dominance region''. In this model, the function $H$ should be minimized with respect to both the sensors location $P$, and the assignment of the dominance regions $\bar{\mathcal{W}}$. A similar deployment model to \eqref{eq::voronoi_opt} has been proposed for multi-sensor deployment with the objective of event detection, particularly when a detailed spatial measurement model for the sensors is unavailable. In such scenarios, it is often customary to assume that due to noise and resolution loss, the sensing performance of a sensor at point $q \in \mathcal{W}$, measured by the $i$-th sensor deployed at location $p_i$, degrades with distance according to a non-decreasing differentiable function $f(\|q - p_i\|)$.

The seminal work by Cortes et al.~\cite{cortes2004coverage} presented a solution to \eqref{eq::voronoi_opt} based on the observation that, at fixed sensor locations, the optimal partition of $\mathcal{W}$ is the Voronoi partition\footnote{We refer to~\cite{okabe2009spatial} for a comprehensive treatment on Voronoi diagrams.} $\mathcal{V}(P) = \{\mathcal{V}_1, \cdots, \mathcal{V}_N\}$ generated by the points $P = \{p_1, \cdots, p_N\}$, where
$$ \mathcal{V}_i = \{ q \in \mathcal{W} \mid \| q - p_i \| \leq \| q - p_j \|, \forall j \neq i \}. $$
Thus, they proposed to write
\begin{align}\label{eq::power_diagram}
    H(P, \bar{\mathcal{W}}) = H_\mathcal{V}(P, \mathcal{V}(P)).
\end{align} Next, by considering the case $f(\|q - p_i\|) = \|q - p_i\|^2$, they proposed a gradient decent flow that can be used to drive a first-order integrator dynamics for each agent $i\in\mathcal{A}$ to a local minimum of~\eqref{eq::voronoi_opt}. This gradient flow continuously moves each agent towards its associated Voronoi centroid. The resulted closed-loop behavior is shown to be adaptive, implementable in a distributed manner by local interaction between Voronoi-neighbor\footnote{Communications structure is specified by the associated Delaunay graph~\cite{cortes2004coverage}.} agents, asynchronous, and provably correct. Including the agents dynamics makes this solution in fact a coverage control problem. 

Later works, such as~\cite{pimenta2008sensing, kwok2010deployment, pimenta2010simultaneous}, sought to model the heterogeneity of agents in their isotropic service capabilities using additively and multiplicatively weighted Voronoi diagrams. In the absence of a detailed model for the agents' service, the heterogeneity of the agents is often represented by power diagrams\footnote{Power diagrams are generalized Voronoi diagrams with additive weights; see~\cite{aurenhammer1987power} for a comprehensive overview.}. Considering an effective range $\boldsymbol{\rho} = \{\rho_1, \cdots, \rho_N\} \subset \mathbb{R}_{>0}$, referred to as the power radius of agents $i \in \mathcal{A}$, in a power diagram model, the service provided by agent $i$ located at $p_i \in \mathcal{W}$ to a point $q \in \mathcal{W}$ is given by the power distance $\|q - p_i\|^2 - \rho_i^2$. Consequently, the area of dominance assigned to the agents is determined by the power diagram $\mathcal{P}(P, \rho) = \{\mathcal{P}_1, \cdots, \mathcal{P}_N\}$, where
\begin{align*}
&\mathcal{P}_i = \{ q \in \mathcal{W} \mid\\
&~\| q - p_i \|^2\! - \rho_i^2 \leq\! \| q - p_j \|^2 \!- \rho_j^2, \forall j \neq i \}. 
\end{align*}
In this scenario, the locational optimization function cost becomes
$$ H_{\mathcal{P}}(P, \boldsymbol{\rho}) = \sum_{i=1}^N \!\int_{\mathcal{P}_i}\!\! (\|q - p_i\|^2 \!- \rho_i^2)  \text{d}(\phi(q)). $$
A simulation scenario is shown in Fig.~\ref{fig::voronoi}, illustrating that agents with a higher power radius $\rho_i$ are assigned a `larger area' to cover. It should be noted that in the special case where $\rho_i = \rho_j$ for all $i, j \in \mathcal{A}$, the power diagram and the Voronoi diagram are identical, i.e., $\mathcal{P}_i = \mathcal{V}_i$. Similar to the homogeneous case, gradient flow solutions have been proposed to determine the final deployment locations of the agents, including approaches that aim to incorporate collision avoidance in the gradient flow dynamics~\cite{arslan2016voronoi}. 

In practice, however, most sensory/service systems, such as cameras, directional antennas, radars, acoustic and ultrasonic sensors are anisotropic. Consequently, attempts have been made to modify Voronoi diagram-based deployment methods to account for multi-agent systems with directional services. For example, \cite{LK-KJ:09}, \cite{FF-ZX-CX-ZT:17}, and \cite{GA-HS-HT-FM:08} consider, respectively, wedge-shaped and elliptic service models and modify the Voronoi diagrams to match the features of the anisotropy of the sensors. Although these extensions incorporate the impact of sensory/service orientation, they often fall short in capturing the detailed physical operation principles if the sensors. Moreover, the sensing/service quality typically does not adhere to a simple (monotonic) functional relation with the Euclidean metric~\cite{BH-NC-KS:13}.
To address this gap, a line of research focused on incorporating detailed sensor models has emerged in the literature. For example,~\cite{arslan2018voronoi, arslan2019statistical, chung2022distributed, ghimire2023optimal, ghimire2024stein} considered coverage problems where the agents' quality of sensing/service is cast as a spatial probabilistic distribution. 
Notably,~\cite{arslan2018voronoi} proposed a specialized form of generalized Voronoi diagrams, termed conic Voronoi diagrams, that considers a visual sensing quality model consistent with the physical nature of cameras.

\begin{figure*}[t]
\centering
\includegraphics[width=0.9\textwidth]{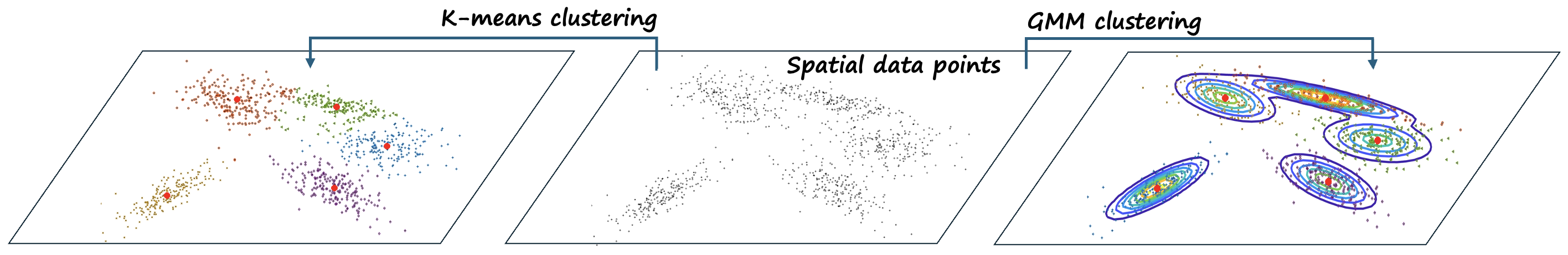}
\caption{Extracting PoIs from $K$-means and GMM clustering. PoIs are shown by filled red dots.}\label{fig::clustering}
\end{figure*}
\section{Static coverage as a discrete task assignment}\label{sec::disc_assign}
In Voronoi-based deployments, the area partitioning and agent assignments occur concurrently. This simultaneous process increases the complexity and may result in a final deployment configuration that does not always place agents in high-density areas, particularly since the solution obtained is a local minimum. The challenge becomes more evident when the number of agents is less than the number of dominant modes in $\phi$, as illustrated in Fig.~\ref{fig::voronoi} (b) and (c) for the deployment of three and two agents when $\phi$ has four dominant modes. As shown in these simulations, in an attempt to provide inclusive coverage, the agents sometimes deploy to points that are equidistant from two or more high-density areas rather than directly within them. While this may be acceptable for facility location applications requiring fair access or dispatch, it is suboptimal for event detection or when the objective is to provide services to points in close proximity to the deployment locations. To address these limitations, the literature has explored deployment strategies that first identify a set of Points of Interest (PoIs) in $\mathcal{W}$, informed by $\phi$, as potential deployment points. These PoIs transform an infinite search space into a finite, representative model of the area. The deployment problem is then solved as a discrete assignment problem, yielding the best coverage objective for the multi-agent team.  
This two-step procedure also provides the flexibility to consider more sophisticated coverage utility measures. Moreover, this approach is well suited for applications such as data harvesting, multi-agent dispatch, and service vehicle deployment in urban areas, where PoIs are predetermined due to operational constraints and there is no flexibility to freely select deployment points in $\mathcal{W}$ based on $\phi$.

In many applications, $\phi$ is derived from the representative spatial data (``point cloud") of the event of interest (targets) over $\mathcal{W}$. A commonly used method for modeling this spatial distribution is the Gaussian Mixture Model (GMM). GMM is a probabilistic model that represents a distribution of data points as a mixture of multiple Gaussian distributions, each with its own mean and variance. The model assumes that data points are generated from several Gaussian distributions, each contributing to the overall probability distribution with some weight~\cite{mcnicholas2016mixture}. By modeling the distribution of the targets, the GMM not only captures this distribution but also intrinsically clusters the targets into subgroups, each represented by a Gaussian basis. In recent work by~\cite{chung2022distributed}, GMM clustering has been used effectively to extract PoIs for deployment.
Alternatively, PoIs can also be identified through various spatial clustering methods using the spatial data themselves. Methods such as $K$-means clustering~\cite{JAK:10} or its variants like Fuzzy C-Means clustering~\cite{JAK:10} can be employed. $K$-means clustering partitions the points into $K \in\mathbb{Z}_{>1}$ clusters, ensuring that each point belongs to the cluster with the nearest mean (cluster center or centroid). An application example in multi-agent deployment can be found in~\cite{ghimire2023optimal,el2020improved}. Figure~\ref{fig::clustering} illustrates the clustering and extraction of PoIs for a set of spatial data points using GMM and K-means methods.

Clustering methods, however, require prior knowledge of the number of clusters and are sensitive to initialization. Recent work in~\cite{ghimire2024stein} proposes an alternative approach to extracting PoIs that circumvents these challenges by using statistical sampling from $\phi$. Specifically,~\cite{ghimire2024stein} suggests utilizing the Stein Variational Gradient Descent (SVGD) method~\cite{QL-JL-MJ:16}. SVGD is a sampling-based deterministic statistical inference method that generates `super samples’ to accurately represent the density distribution $\phi$. SVGD is known for its effectiveness even with a small number of samples, making it highly suited for extracting PoIs. However, sampling-based algorithms face the risk of generating samples too close to each other, leading to overlapping coverage by deployed agents.
In light of this observation,~\cite{ghimire2024stein} carefully designs the sample-spread mechanism of the SVGD algorithm to generate samples that are not only representative of the distribution $\phi$ but also appropriately spread according to the effective service/sensing footprint of the agents.

\begin{figure}[H]
\begin{center}
\includegraphics[width=0.33
\textwidth]{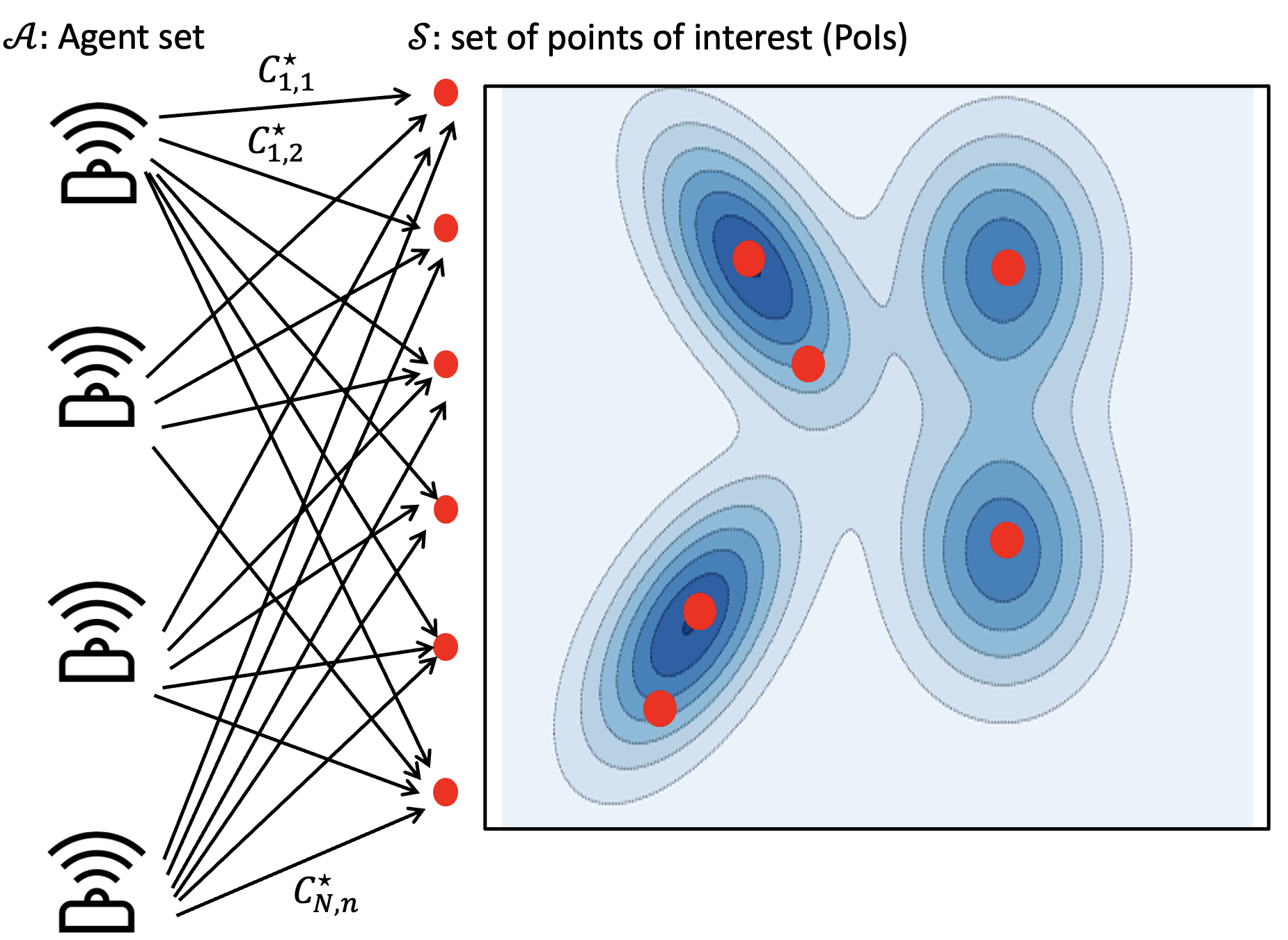}
\caption{Schematic illustration of the multi-agent assignment as an optimal bipartite matching problem, where red dots denote PoIs. }\label{fig::matching}
\end{center}
\end{figure}

Nevertheless, given a finite set of PoIs denoted by $\mathcal{S}=\{1,\cdots,n\}$, as depicted in Fig.~\ref{fig::matching}, a natural solution for agent assignment is to use optimal bipartite matching through the optimal linear assignment problem, also known as the discrete optimal mass transport problem. This problem can be formulated as follows:
\begin{subequations}
\label{eq::assignment}
\begin{align}
&\!\!\!\boldsymbol{Z}^\star= \arg\min\sum\nolimits_{j\in\mathcal{S}}\!\sum\nolimits_{i\in\mathcal{A}}\!\!\!Z_{i,j}C_{i,j}^\star,\\
    & Z_{i,j}\in\{0,1\},\quad i\in\mathcal{A},~j\in\mathcal{S},\\
     & \sum\nolimits_{j\in\mathcal{S}}\!\!\!Z_{i,j}=1,~ \forall i\in \mathcal{A},\label{eq::assignment_sensor} \\ &\sum\nolimits_{i\in\mathcal{A}}\!\!\!Z_{i,j}\leq 1, ~ \forall j\in\mathcal{S}.\label{eq::assignment_cluster}
\end{align}
\end{subequations}
where $Z_{i,j}$ is the assignment indicator with $Z_{i,j}=1$ if the $i$-th agent is assigned to the $j$-th PoI and $Z_{i,j}=0$ otherwise, and $C_{i,j}^\star$ is the minimum cost of assigning the $i$-th agent to the $j$-th PoI. With the assumption that $|\mathcal{A}| < |\mathcal{S}|$, constraint~\eqref{eq::assignment_sensor} ensures that every agent gets assigned to one PoI, and constraint~\eqref{eq::assignment_cluster} ensures that every PoI is  assigned at most to one agent.  The optimization problem~\eqref{eq::assignment} is a standard assignment problem and can be solved using existing algorithms such as the Hungarian algorithm~\cite{KW-55}. It can also be solved through linear programming via a continuous relaxation approach, even in a distributed manner using, for example, the distributed simplex algorithm proposed by~\cite{BM-NG-BF-AF:12}. The solution of the assignment problem~\eqref{eq::assignment} gives the final deployment configuration of the agents.

Considering an effective footprint $\mathcal{C}_i(p_i,\theta_i)=\{q\in\mathcal{W}\mid F_i(q|p_i,\theta_i)\leq 0\}\subset\mathcal{W}$, a compact set, for each agent $i\in\mathcal{A}$ deployed at position $p_i \in \mathcal{W}$ and orientation $\theta_i \in [0, 2\pi]$, the assignment framework provides flexibility in computing the deployment cost $C_{i,j}^\star$ via various measures such as
\begin{align*} C_{i,j}^\star = \min_{\theta \in \Theta} \int_{\mathcal{C}_i(p_j,\theta)} \!f_i(\|q - p_j\|) \text{d}\phi(q), \end{align*}
or as proposed in ~\cite{ghimire2024stein} as
\begin{align*} C_{i,j}^\star = \min_{\theta \in \Theta} \{\mathcal{K}\mathcal{L}(s_i(q|&p_j,\theta)|\phi(q)) ~~\text{for}\\& ~~~q \in \mathcal{C}_i(p_j,\theta)\}, \end{align*}
where the Kullback–Leibler divergence (KLD)\footnote{ Given two continuous probability density distributions $\psi(\boldsymbol{x})$ and $\phi(\boldsymbol{x})$, $\boldsymbol{x} \in \mathbb{X}$, KLD is defined as $\mathcal{K}\mathcal{L}\big(\psi(\boldsymbol{x})||\phi(\boldsymbol{x})\big) = \int_{\boldsymbol{x} \in \mathbb{X}} \psi(\boldsymbol{x}) \log \frac{\psi(\boldsymbol{x})}{\phi(\boldsymbol{x})}\text{d}\boldsymbol{x},$ which is a measure of similarity (dissimilarity) between the two probability distributions; the smaller the value, the more similar the two distributions are. KLD is zero if and only if the two distributions are identical~\cite{MDJC:03}. }, denoted by $\mathcal{K}\mathcal{L}$, is used to measure the similarity between the $i$-th agent's spatial service, cast as a probability distribution $s_i(q|p_j,\theta)$ when it is located at $p_j$ with orientation $\theta$, and $\phi(q)$ over the effective footprint $\mathcal{C}_i(p_j,\theta)$ of the agent. In computing $C_{i,j}^\star$ for directional agents, the models above search over a finite number of deployment orientations for each agent, that is, $\theta \in \Theta = \{\bar{\theta}_1, \cdots, \bar{\theta}_M\} \subset [0, 2\pi]$. 

For the special case of a Gaussian service distribution $s_i(q|p,\theta)=\mathcal{N}(q|p,\bar{\Sigma}_i(\theta))$ and PoIs extracted from a GMM clustering method,~\cite{chung2022distributed} defined $C_{i,j}^\star$ as the weighted KLD difference between the agent's service distribution $s_i$ and the $j$-th Gaussian basis of the GMM-modeled area priority function $\phi(q) = \sum_{j=1}^n \pi_j \mathcal{N}(q|p_j, \Sigma_j)$. They showed that the optimal deployment for agent $i$ in cluster $j$ is to align the means of $s_i$ and $\mathcal{N}(q|p_j, \Sigma_j)$, i.e., deploy agent $i$ at $p_j$, and make the principal axis of $s_i$ parallel to that of $\Sigma_j$. Because the distributions compared are Gaussian,~\cite{chung2022distributed} was able to compute $C_{i,j}^\star$ in closed form without the need to search over $\Theta$ for the best deployment orientation. Alternatively,~\cite{ghimire2023optimal} computed $C_{i,j}^\star$ by sampling from $s_i$ and using a discrete optimal mass transport method to measure the statistical distance between the samples drawn from $s_i$ and the target points in the $j$-th cluster created by the $K$-means method. They used a modified version of the discrete optimal mass transport where the rotation and translation of the point cloud samples drawn from $s_i$ are part of decision variables. This method, inspired by the Iterative Closest Point (ICP) algorithm used in point-set registration in computer vision~\cite{JY-HL-DC-YJ:16}, allows for the computation of $C_{i,j}^\star$ at the best deployment position and orientation.

Despite the flexibility and tractability offered by the linear optimal assignment optimization in~\eqref{eq::assignment}, this formulation does not account for the consequences of overlapping coverage. As an alternative, the PoI-assignment problem--assigning $N$ elements from the PoIs set $\mathcal{S} = \{1, \cdots, n\}$ to $N$ agents $\mathcal{A} = \{1, \cdots, N\}$--can be formulated as a set function maximization problem. For homogeneous agents, the set function maximization problem is expressed~as: 
\begin{align}\label{eq::submdoular_uniform}
    \mathcal{R}^\star =&\arg\max_{\mathcal{R}\subset\mathcal{S}} f(\mathcal{R})~~\text{subject to~}\\& |\mathcal{R}|\leq N,\nonumber
\end{align}
where $f(\mathcal{R}): 2^\mathcal{S}\to\mathbb{R}_{\geq0}$ represents the joint utility of deploying the agents at $\mathcal{R} \subset \mathcal{S}$. For heterogeneous agents, the set function maximization problem is given by: 
\begin{align}\label{eq::submdoular_partition}
    \mathcal{R}^\star =&\arg\max_{\mathcal{R}\subset\bar{\mathcal{S}}} f(\mathcal{R})~~\text{subject to~} \\
    &|\mathcal{R}\cap \mathcal{S}_i|\leq 1, ~i\in\mathcal{A},\nonumber
\end{align}where $\mathcal{S}_i = \{ (i, j) \mid j \in \mathcal{S} \}$ and $\bar{\mathcal{S}}=\cup_{i\in\mathcal{A}} \mathcal{S}_i$. The constraint in~\eqref{eq::submdoular_partition} ensures that each agent is assigned only one PoI.

Combinatorial optimization problems of the form~\eqref{eq::submdoular_uniform} and~\eqref{eq::submdoular_partition} are often NP-hard. However, for a special class of set functions known as \emph{submodular} functions, the seminal work by Nemhauser, Wolsey, and Fisher in the 1970s~\cite{LMF-GLN-LAW:78,GLN-LAW-MLF:78,GLN-LAW:78} showed that the so-called \emph{sequential greedy algorithm} can deliver a suboptimal solution with a well-defined optimality gap in polynomial time. When the utility function is submodular, the optimization problem~\eqref{eq::submdoular_uniform} is referred to as submodular maximization subject to a uniform matroid. In this case, the sequential greedy algorithm starts with $\mathcal{R}_{\text{SG}} = \emptyset$ and iterates according to \begin{align*}
&p^\star=\arg\!\!\!\!\max_{p\in\mathcal{S}\backslash\mathcal{R}_{\text{SG}}}\!\! (f(\mathcal{R}_{\text{SG}}\cup\{p\})-f(\mathcal{R}_{\text{SG}}))\\
&    \mathcal{R}_{\text{SG}}\leftarrow\mathcal{R}_{\text{SG}}\cup \{p^\star\}
\end{align*} until $N$ elements are selected. Alternatively, the optimization problem~\eqref{eq::submdoular_partition} is referred to as submodular maximization subject to a partition matroid. In this case, the sequential greedy algorithm starts with $\mathcal{R}_{\text{SG}} = \emptyset$ and iterates according to
\begin{align*}
&p^\star=\arg\max_{p\in\mathcal{S}_i} (f(\mathcal{R}_{\text{SG}}\cup\{p\})-f(\mathcal{R}_{\text{SG}}))\\
&    \mathcal{R}_{\text{SG}}\leftarrow\mathcal{R}_{\text{SG}}\cup \{p^\star\}
\end{align*}
until $i = N$ and each agent is assigned a PoI.

Many coverage utility functions, such as max-cover, facility location, and mutual information functions, are known to be submodular~\cite{AK-DG:14bookChap}. Several well-known submodular maximization frameworks can also be applied to coverage problems. For example, consider the \emph{Exemplar-based Clustering} method introduced by~\cite{LK-PR:09}, which aims to identify a subset of exemplars that optimally represent a large dataset by solving the $K$-medoid problem. This method minimizes the cumulative pairwise dissimilarities between chosen exemplars $\mathcal{S}$ and dataset elements $\mathcal{D}$: \begin{equation} \label{eq::util_min} L(\mathcal{R}) = \sum_{p \in \mathcal{R}} \min_{d \in \mathcal{D}} \text{dist}(p, d), \end{equation} for any subset $\mathcal{R} \subset \mathcal{S}$, where $\text{dist}(p, d) \geq 0$ defines the dissimilarity, or distance, between elements. To find an optimal subset $\mathcal{R}$ that minimizes $L$, this problem is posed as a submodular maximization problem with the utility function: \begin{align} \label{eq::util_num} f(\mathcal{R}) = L(d_0) - L(\mathcal{R} \cup d_0), \end{align} where $d_0$ is a hypothetical auxiliary element. This utility function quantifies the reduction in loss from the active set versus using only the auxiliary element and is submodular and monotonically increasing~\cite{RG-AK:10}.
An instance of using exemplar-based clustering for multi-agent deployment is simulated in~\cite{NR-SSK:23}. The problem considered is an information harvesting task that aims to collect data from sources $\mathcal{D} \subset \mathcal{W}$. The goal is to deploy $N$ data harvesters at pre-specified points $\mathcal{S}$. In this problem, the number of agents is far less than the number of PoIs, i.e., $N < |\mathcal{S}|$. The optimal deployment minimizes the distance between each information point $d \in \mathcal{D}$ and its nearest harvester at $b \in \mathcal{S}$. ~\cite{NR-SSK:23} formulates this problem as an exemplar clustering problem using the submodular utility function~\eqref{eq::util_num}, with $\text{dist}(b, d) = \|b - d\|$ representing the Euclidean distance. For a numerical demonstration, see~\cite{NR-SSK:23}.

\section{Continuum task assignment for large-scale swarms}\label{sec::continum}

\begin{figure*}[ht!]
\centering
\includegraphics[width=0.7\textwidth]{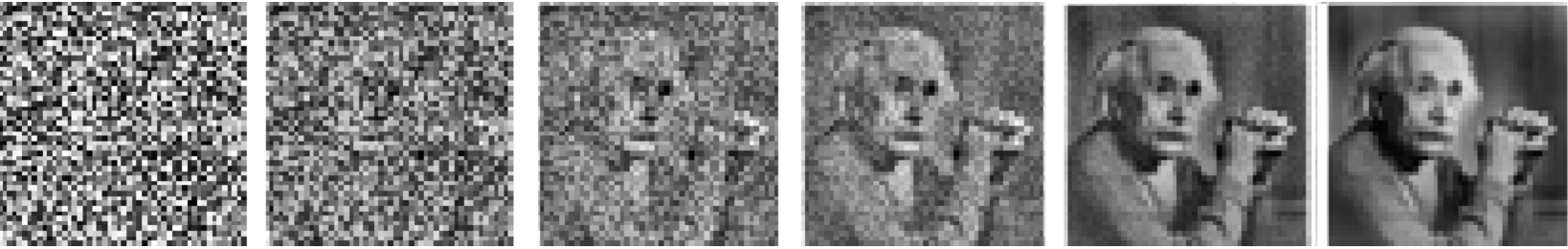}
\caption{A swarm of agents evolve under the distributed transport algorithm of~\cite{VK-SM:18-cdc} to reconfigure into an image. Figure courtesy of Vishaal Krishnan.}
\label{einstein}
\end{figure*}

In spatial coverage control problems that involve a very large number of agents, it is more meaningful to specify both the task assignment and swarm control objectives in a macroscopic manner, as a distribution matching problem, whereby the distribution of agents is to coincide with that generating the tasks. 

In fact, one may ask if such objective could be achieved by taking the limit on the number of agents to infinity in the locational optimization problem formulations of the previous section. At the same time, one may think that the problems of local optima, which result from a poor initialization of Lloyd's algorithm, may be addressed in this way. An answer to these questions can be provided via the so-called Wasserstein-Kantorovich metric, which solves the problem of optimal transport, also called the Earth Mover's problem.

Formally, given two measures $\mu,\nu$ over a space $\mathcal{W}$ with bounded $p$ moments, the $p$-Wasserstein-Kantorovich metric $W_p$ is defined as 
\begin{align*}&W_p^p(\mu,\nu) =\\&~~\qquad \min_{\pi\in \Pi(\mu,\nu)} \int_{\mathcal{W} \times \mathcal{W}} c(x,y)^p \text{d}\pi(x,y),\end{align*} where $c$ is the cost of transport from $x$ to $y\in \mathcal{W}$ and $\Pi(\mu,\nu)$ is the set of measures over $\mathcal{W}\times \mathcal{W}$ with marginals $\mu$ and $\nu$, respectively. The Wasserstein metric finds many applications, and directly generalizes the solution of a discrete optimal task assignment problem 

It turns out that the most basic Expected Value metric (locational optimization cost) for coverage control is equal to the best optimal transport from an agent-based discrete distribution to the target distribution of tasks, $\text{d}\phi$~\cite{VK-SM:22}. Specifically, such discrete distribution is naturally defined via the Voronoi partition  $\mathcal{V} = \{\mathcal{V}_1,\dots,\mathcal{V}_N\}$ generated by the agent locations $p_1,\dots, p_N$, and  $\nu = \text{d}\phi(q)$. To define it, let $w_i$ be the mass of $\text{d}\phi(q)$ over the Voronoi region $\mathcal{V}_i$, and $\delta_{p_i}$ be the delta distribution over $p_i$, for $i \in\{ 1,\dots,N\}$. Then, the associated discrete measure is $\mu_{\mathcal{V}} = \sum_{i=1}^N w_i \delta_{p_i}(q)$, assigning all the mass $w_i$ to the location of agent $i$, $p_i$.  It is not hard to show that that $W_2(\mu_{\mathcal{V}},\text{d}\phi) = \mathcal{H}_{\mathcal{V}}(P)$. This property sheds light on the both questions above: first, at very large scales, when the number of agents goes to infinity, the cost $\mathcal{H}_{\mathcal{V}}(P)$ approaches zero. Thus, in the limit, the original $\mathcal{H}$ can be brought down to 0 just by adding (uniformly at random) more and more agents, regardless of their positions inside their Voronoi regions. Thus, the most basic locational optimization cost does not lead to the large-scale goal of distribution matching. As discussed in~\cite{VK-SM:22}, the utilization of \textit{generalized Voronoi partitions} as in~\cite{kwok2010deployment} that are also \textit{equitable}; that is, for which region masses are identical; $w_i = w_j$ for all $i,j in\{ 1,\dots, N\}$, solves this problem. Under this constraint, we do have consistency of problems and objectives. 

The Wasserstein metric has additional convexity properties that can help us answer the question of optimality as well: while the finite-agent locational optimization problems are non-convex, $W_2$ enjoys generalized convexity (in sense of the so-called displacement interpolations, and convexity with respect to a target measure), which allows us to roughly state that, by taking the limit in the number of agents to infinity, we are ``making the problem convex''. 

However, while this is satisfactory theoretically, the question of how to devise new and tractable algorithms that can be still applicable for a large, but finite number of agents still remains. As actuation rests at the microscopic scale and the individual agents, it is still necessary to ensure that the obtained algorithms satisfy the microscopic constraints that make them implementable, such as that of limited sensing, communication, computation, asynchronous interactions, and distributed control.  

One can look for an answer to this question by considering modern, discrete-time optimization techniques that, at the macro-scale, solve the desired optimization problem, and that, at the micro-scale, behave adequately for discrete multi-agent systems. In~\cite{VK-SM:22, VK-SM:18-cdc} we propose an algorithmic approach  that exploits proximal gradient optimization together with variational problem discretization via sampling. In particular, the proposed algorithms retain the distributed implementation properties required by scalable multi-agent coordination algorithms.

Figure~\ref{einstein} illustrates an implementation of one of such algorithms for the optimal transport to a target distribution. The target distribution is a pixelated image of Einstein (higher intensity represents higher number of agents at that location). The algorithm allows agents to compute the global optimal optimal transport to the target density in a distributed manner, in the sense of the graph induced by the Voronoi partition. The cost of transport $c(x,y)$ is given by the Euclidean distance.

\section{Concluding Remarks}\label{sec::conclude}

The multi-agent spatial coverage problem is a fundamental challenge in robotics, with wide-ranging applications that demand innovative and adaptive solutions. Throughout this note, we have provided a quick exposition to various methodologies that address its various aspects, providing insights into different strategies and their applicability to specific scenarios. 

While this note is by no means a comprehensive overview of the vast literature on multi-agent spatial coverage, it aims to showcase some solution approaches and illustrate how different problem settings and operational assumptions can significantly influence the solution strategies we develop. By covering a wide range of methodologies, from classic Voronoi-based deployments to modern statistical sampling techniques, and exploring both deployment problems with a limited number of agents and a continuum of agents, we hope to provide a useful perspective on this complex and evolving field.

The multi-agent spatial coverage problem continues to be a vibrant area of research with numerous practical applications, ranging from environmental monitoring and data harvesting to service vehicle deployment and large-scale swarm behavior. The diverse approaches discussed in this paper highlight the flexibility and adaptability required to tackle various challenges inherent in different operational contexts.

Future research in this area is likely to benefit from advances in machine learning, optimization algorithms, and decentralized control methods. Embracing these theories can lead to more efficient, scalable, and robust solutions. Moreover, addressing open challenges, such as dealing with overlapping coverage, optimizing deployment in heterogeneous environments, and developing methods that can adapt to dynamic changes in real time, will further enhance the effectiveness of multi-agent systems.

Ultimately, we hope this note will serve as an introduction and a reference point for researchers and practitioners, encouraging them to explore the rich and diverse methodologies available for solving multi-agent spatial coverage problems. By understanding the impact of different problem settings and operational assumptions, we can develop more tailored and effective solutions, advancing the state-of-the-art in this fascinating field.

\section{Bios}
\includegraphics[width=0.3
\textwidth]{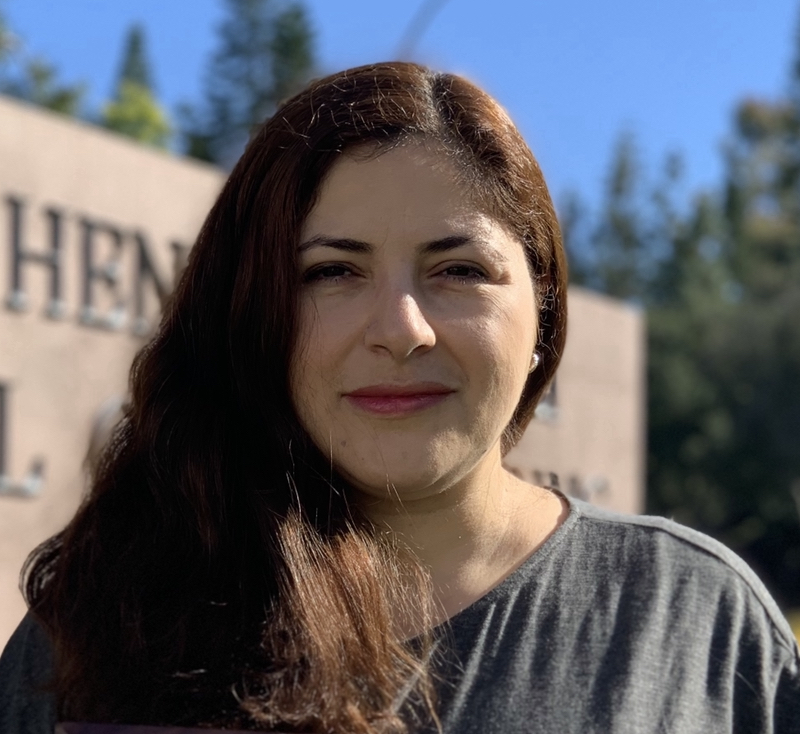}
\textbf{Solmaz Kia} is a Professor of Mechanical and Aerospace Engineering at the University of California, Irvine (UCI), CA, USA. She also holds a joint appointment in the Computer Science Department at UCI. She obtained her Ph.D. in Mechanical and Aerospace Engineering from UCI in 2009, and her M.Sc. and B.Sc. in Aerospace Engineering from Sharif University of Technology, Iran, in 2004 and 2001, respectively. From June 2009 to September 2010, she was a Senior Research Engineer at SySense Inc., El Segundo, CA. She held postdoctoral positions in the Department of Mechanical and Aerospace Engineering at the University of California, San Diego, and UCI.
She was a recipient of the UC President’s Postdoctoral Fellowship in 2012-2014, an NSF CAREER Award in 2017, and the Best Control System Magazine Paper Award in 2021. She is a Senior Member of IEEE. She serves as an Associate Editor for Automatica, IEEE Transactions on Control of Network Systems, and IEEE Open Journal of Control Systems. Her main research interests include distributed optimization, coordination, estimation, nonlinear control theory, and probabilistic robotics navigation and motion planning (email: solmaz@uci.edu).

\includegraphics[width=0.3\textwidth]{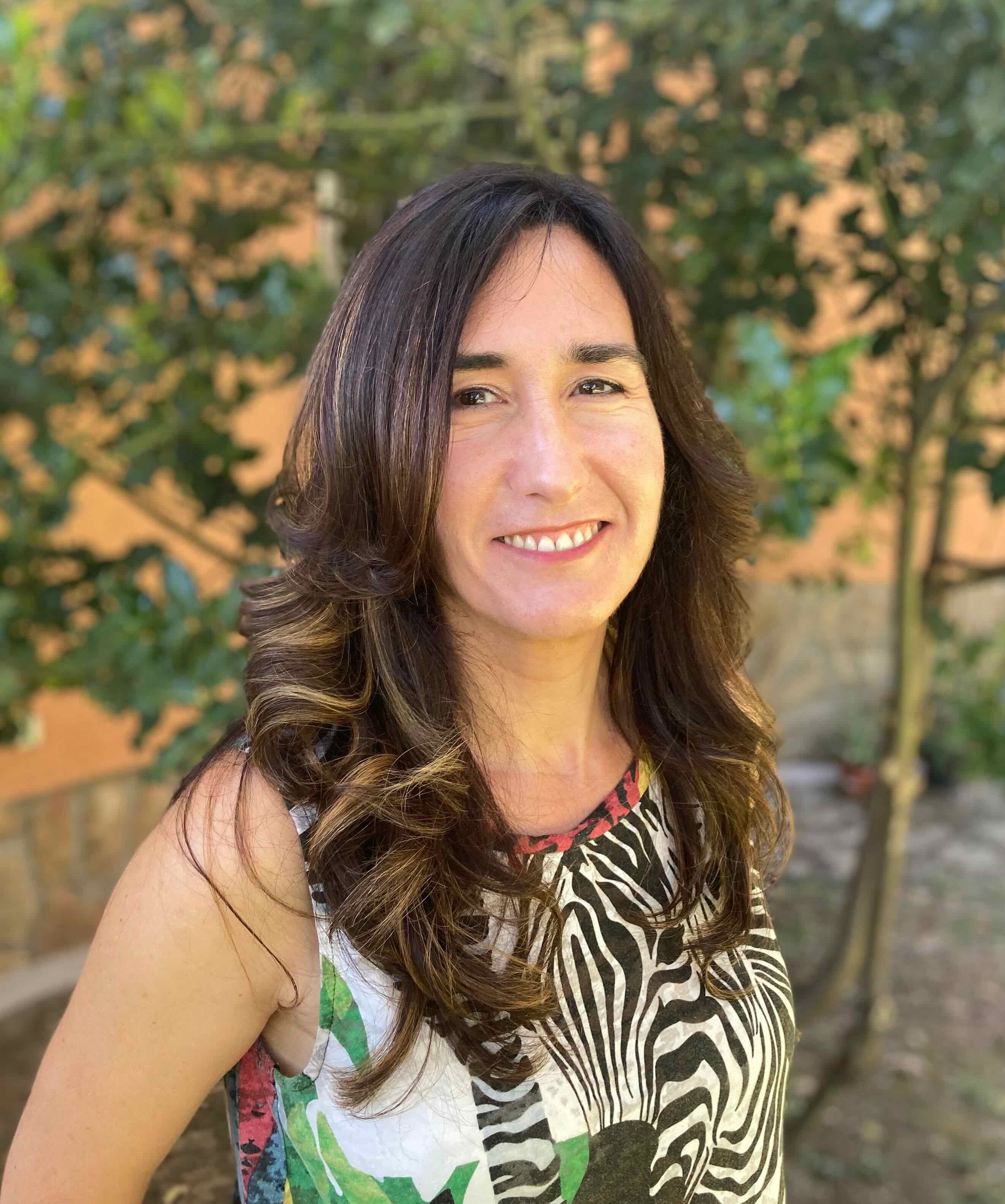}
\textbf{Sonia Martinez} Sonia Martinez is a Professor of Mechanical and Aerospace Engineering at the University of California, San Diego, CA, USA. She received her Ph.D. degree in Engineering Mathematics from the Universidad Carlos
III de Madrid, Spain, in May 2002. She was a Visiting Assistant
Professor of Applied Mathematics at the Technical University of
Catalonia, Spain (2002- 2003), a Postdoctoral Fulbright Fellow at the
Coordinated Science Laboratory of the University of Illinois,
Urbana-Champaign (2003-2004) and the Center for Control, Dynamical
systems and Computation of the University of California, Santa Barbara
(2004-2005). Her research interests include the control of networked
systems, multi-agent systems, nonlinear control theory, and planning
algorithms in robotics. She became a Fellow of IEEE in 2018. She is a
co-author (together with F. Bullo and J. Cortés) of ‘‘Distributed
Control of Robotic Networks’’ (Princeton University Press, 2009). She
is a co-author (together with M. Zhu) of ‘‘Distributed
Optimization-based Control of Multi-agent Networks in Complex
Environments’’ (Springer, 2015). She is the Editor in Chief of the
recently launched CSS IEEE Open Journal of Control Systems, and Senior Editor for Surveys of Automatica. (email: soniamd@ucsd.edu)

\bibliographystyle{plain}

\end{multicols}

\end{document}